\pdfoutput=1

\documentclass[11pt]{article}

\usepackage[]{acl}
\usepackage{times}
\usepackage{latexsym}

\usepackage{relsize}
\usepackage{graphicx}
\graphicspath{ {./img/} }
\usepackage{tcolorbox}
\usepackage[flushleft]{threeparttable}
\usepackage{url}
\usepackage{hyperref}
\usepackage{xstring}
\usepackage{xspace}
\usepackage{covington}
\usepackage[shortlabels]{enumitem}
\usepackage{booktabs}
\usepackage{multirow}
\usepackage{xcolor}
\usepackage{soul}
\usepackage{listings}
\usepackage{subcaption}
\usepackage{gb4e}
\noautomath

\definecolor{lightblue}{RGB}{200,245,255}
\definecolor{lightred}{RGB}{255,200,200}
\definecolor{lightgreen}{RGB}{205,255,200}
\definecolor{lightgrey}{RGB}{225,225,225}

\newcommand{\hllb}[2][lightblue]{{\sethlcolor{#1}\hl{#2}}}
\newcommand{\hllr}[2][lightred]{{\sethlcolor{#1}\hl{#2}}}

\newcommand{\nrt}{{NoReC\textsubscript{\textit{tsa}}}\xspace}
\newcommand{\nrf}{{NoReC\textsubscript{\textit{fine}}}\xspace}
\newcommand{\fen}{F\textsubscript{1}\xspace}

\newcommand{\norex}[1]{\textit{#1}}
\newcommand{\eng}[1]{`#1'}

\usepackage[T1]{fontenc}

\usepackage[utf8]{inputenc}

\usepackage{microtype}

\title{Entity-Level Sentiment Analysis (ELSA):\\ An exploratory task survey}

  \author{Egil Rønningstad \and Erik Velldal \and Lilja {\O}vrelid  \\
  University of Oslo\\
  Department of Informatics\\
 {\tt \{egilron,erikve,liljao\}@ifi.uio.no}}

\begin{document}
\maketitle
\begin{abstract}
This paper explores the task of identifying the overall sentiment  expressed towards volitional entities (persons and organizations) in a document -- what we refer to as Entity-Level Sentiment Analysis (ELSA). While identifying sentiment conveyed towards an entity is well researched for shorter texts like tweets, we find little to no research on this specific task for longer texts with multiple mentions and opinions towards the same entity.  This lack of research would be understandable if ELSA can be derived from existing tasks and models. To assess this, we annotate a set of professional reviews for their overall sentiment towards each volitional entity in the text. We sample from data already annotated for document-level, sentence-level, and target-level sentiment in a multi-domain review corpus, and our results indicate that there is no single proxy task that provides this overall sentiment we seek for the entities at a satisfactory level of performance. We present a suite of experiments aiming to assess the contribution towards ELSA provided by document-, sentence-, and target-level sentiment analysis, and provide a discussion of their shortcomings. We show that sentiment in our dataset is expressed not only with an entity mention as target, but also towards targets with a sentiment-relevant relation to a volitional entity. In our data, these relations extend beyond anaphoric coreference resolution, and our findings call for further research of the topic. Finally, we also present a survey of previous relevant work. 
\end{abstract}

\section{Introduction}

Over the course of the last two decades, the field of NLP has generated a vast body of research on sentiment analysis (SA), i.e. the task of identifying opinions expressed in text. Prior work has focused on a range of different levels of analysis; from document- or sentence-level polarity classification to more fine-grained prediction of various components of opinions, like source/holder expressions, polar expressions, target expressions and aspect-based sentiment classification.

However, we observe that a more aggregated level of analysis, what we here dub "entity-level sentiment analysis" (or ELSA for short), remains under-explored. For our purposes, we will define ELSA as the task of determining the overall (i.e. document-level) polarity (positive/negative) expressed towards an entity in a text. Moreover, for the current paper we will restrict the discussion to volitional entities, like persons (PER) or organizations (ORG).

A given text might make reference to multiple distinct entities, each of which might be mentioned multiple times, both directly and indirectly, and also have multiple opinions directed towards them. Hence, solving the task of ELSA in its full complexity may potentially involve several different sub-tasks, including (but not necessarily limited to) named entity recognition (NER), resolution of entity mentions, coreference- and anaphora-resolution, identification of sentiment targets and/or aspects and their polarities, and finally what we here refer to as "target--entity resolution", i.e. the task of identifying the particular entity with which a given target expression is associated.

The main goal of this paper is to shed more light on this task of entity-level sentiment analysis. To better understand the complexity of the task, we quantify how far we can potentially get toward the goal of ELSA by simply building on existing tools for SA at lower levels. We do this through an exploratory analysis of an existing SA dataset that comprises annotations for several levels of granularity, for which we experiment with different strategies for aggregating these gold annotations to infer entity-level sentiment. Importantly, we also discuss which pieces appear to currently be missing in order to fully solve ELSA. We start, however, by surveying relevant prior work, and also discuss the often diverging terminology used in the field. 

The paper is structured as follows. 
In Section~\ref{sec:literature_review} we first present background literature on the task of analyzing sentiment towards individual entity representations, before we discuss the limited work we found on resolving sentiment-relevant elements to the document-level.  We found no previous work describing the ELSA task as we defined it, and very little in general on resolving sentiment from the sub-sentence level to the document-level. 
Section~\ref{sec:dataset} presents the datasets we sample from; Norwegian data annotated for document-level, sentence-level and target-level sentiment classification. We also present our new, exploratory dataset, annotated directly for document-level sentiment classification for each volitional entity in the text. In Section~\ref{sec:analysis_doc_sent} we present our results from attempting to derive ELSA from document-level and sentence-level sentiment analysis -- what we dub target-independent approaches. We find that merely locating an entity mention inside a positive document is a very weak indicator of a positive sentiment towards that entity.  Section~\ref{sec:analysis_target_dep} presents our findings from deriving ELSA classification from Targeted Sentiment Analysis (TSA). Besides discrepancies from annotator disagreement, we find that in order to fully solve the ELSA task, future work should aim to add another level of analysis, corresponding to the relations between sentiment targets and their affiliated entities. In  Section~\ref{sec:modeling} we report on a baseline model for ELSA through TSA as a proxy task. 

\section{Literature review}\label{sec:literature_review}
This section first presents prior work related to ELSA. We then survey the terminology used in the literature, comparing this to our suggested definition of entity-level SA.

 \subsection{Related work on short texts} \label{sec:related_short}

\citet{mitchell-etal-2013-open} introduce the task of identifying which occurrences of named entities are sentiment targets in a text, and further to classify the sentiment towards these as positive or negative. They narrow the scope of named entities to volitional entities; i.e.  organizations and persons. 
\citet{zhang-etal-2015-neural} follow up on this work and expand on the TSA task description. 

The work of \citet{mitchell-etal-2013-open} is highly related to our task in that they for each text identify the sentiment polarity towards each volitional entity mentioned in the text. However, while \citet{mitchell-etal-2013-open} work on Twitter data, hence by their nature very short texts, our goal in this work is to extend the task to longer texts where the sentiment towards one volitional entity may be expressed through multiple mentions and opinion expressions. 

On a similar note, \citet{jiang-etal-2011-target} also analyze sentiment towards each target in tweets. The goal of their work is, for a corpus of tweets and a query term, to return tweets classified as positive or negative towards the entity in the query term. The work goes beyond situations where the named entity is the sentiment target, into what is described as "extended targets". This work is a step in the direction of locating an extended set of segments of the text that are relevant for the sentiment towards each entity.

 \subsection{Related work on longer texts}\label{sec:related_long}
 One possible way of aggregating sentiment expressions is through the application of Coreference Resolution (CR) techniques.
\citet{de-clercq-hoste-2020-absolutely} explore the benefits of CR in Aspect-Based Sentiment Analysis (ABSA). They find that when adding gold-standard coreference information, their models increase their performance by up to one percentage point, while using automatically retrieved CR data \citep{lee2013deterministic, de-clercq-etal-2011-cross} would have no, or even negative effect on the performance of their ABSA models.  
\citet{stoyanov-cardie-2006-partially} explore the benefits of holder coreference resolution as a method for extracting the document-level sentiment expressed by one holder. The task of target coreference resolution is mentioned in suggestions for future work.
\citet{farra-mckeown-2017-smarties} address the task of open-domain TSA, where their goal is to cluster targets and identify salient entities towards which opinions are expressed in the text. These targets may be nouns, noun phrases, events or concepts.

\citet{steinberger-etal-2017-large} present their Europe Media Monitor (EMM) system, where the overall task is to detect the positive or negative sentiment towards persons and organizations. They process about 70 languages and therefore use linguistically light-weight methods that can work with low-resource languages.

The recent work of \citet{https://doi.org/10.48550/arxiv.2204.06893} addresses some of the same limitations in previous work as observed by us, namely the limited focus on identifying sentiment targets only at the level of sentences or tweets. A new multi-domain dataset is provided, annotated for sentiment targets and polarity, where all targets referring to the same entity are joined in a nested target structure. Their dataset provides a step towards targeted sentiment analysis at the document-level. 
There is no special treatment of volitional entities however.
 


\subsection{Terminology review}\label{sec:terminology_review}

Of the few papers we could find that use the term "entity-level sentiment analysis", several of them appear  to treat it as synonymous to targeted SA (TSA)  \cite{li2017learning, alimova-tutubalina-2019-entity, huang2020entity,sweeney-padmanabhan-2017-multi, engonopoulos2011els}. In TSA, the task is for each sentence to extract any segment being the target of a sentiment expression, and the sentiment polarity towards this target. This notion of a  target is in line with the target expressions in widely used datasets for TSA \citep{pontiki-etal-2014-semeval,pontiki-etal-2016-semeval}, where there is no linking or aggregation to the document-level. Having the TSA term to describe this task, we suggest that the ELSA term is a better fit for an aggregated entity-level, where one entity may be linked to several targets in multiple sentences.

We did, however, also find a few studies employing the term "entity-level sentiment analysis" about entities aggregated from several mentions in the text \citep{farra-mckeown-2017-smarties, steinberger-etal-2017-large, https://doi.org/10.48550/arxiv.2204.06893}. These papers were presented in Section~\ref{sec:related_short}.

\subsection{Conclusions from the literature study} 

We find that while there has been quite some work on classifying the polarity of particular occurrences of named entity mentions in text, like that of \citet{mitchell-etal-2013-open} and \citet{zhang-etal-2015-neural}, we have seen only a few studies that attempt to link or cluster several related segments of a document to resolve sentiment towards entities at the document-level. From existing research, we have seen that coreference resolution \cite{stoyanov-cardie-2006-partially}, PMI \cite{jiang-etal-2011-target} and semantic clustering \cite{farra-mckeown-2017-smarties} have been used to resolve which entity mentions belong together. 
None of these papers were found reporting on actual experiments for the task of assigning one sentiment classification per volitional entity per document though. The recent paper by \citet{https://doi.org/10.48550/arxiv.2204.06893} represents our closest match.


While we find that our usage of the term "entity-level sentiment analysis" is thematically related to a few other usages in the literature, we do not see any established competing use of the term. We therefore suggest ELSA as an appropriate and descriptive term for the task discussed in the current paper. 


\section{The sentiment dataset} \label{sec:dataset}
In order to investigate  how the ELSA task relates to pre-existing sentiment analysis tasks, we wanted to build on an existing full-text document collection that is annotated for sentiment information at several levels of analysis. The suite of annotated datasets that are based on the Norwegian Review Corpus -- NoReC \citep{velldal-etal-2018-norec} -- fits this bill. While  Section~\ref{sec:exploratory} describes how we build on NoReC to create an exploratory dataset for ELSA, we first describe the different levels of existing annotations in NoReC below.  

\subsection{NoReC}\label{sec:norec} 
NoReC is a multi-domain dataset of full-text professional reviews published in Norwegian online news sources, and a subset of the documents have been annotated for fine-grained and sentence-level sentiment. Each review in NoReC is accompanied by a rating given by the reviewer, on a scale from 1 to 6.  We here take this to serve as a polarity label for the overall document.

\paragraph{Fine-grained sentiment}  
A subset of NoReC has been annotated for fine-grained  sentiment information in \nrf \citep{ovrelid-etal-2020-fine}, including holders, target expressions, polar expressions, polarities, and polar intensities.  This was one of the datasets used in the recent SemEval shared task on structured sentiment analysis -- SemEval-2022 Task 10 \citep{barnes-etal-2022-semeval}.\footnote{\url{https://competitions.codalab.org/competitions/33556}} 
As presented in  Table~\ref{table:norec_fine_stats_rate}, the \nrf training split consists of 327 documents, comprising 8634 sentences, giving an average of 26.4 sentences per document. The training split contains 5000 unique sentiment targets. Based on this data we can derive datasets for TSA as well as sentence-level SA, as described below.

\paragraph{Target-level sentiment} 
We here describe how we derive a dataset for targeted SA -- dubbed \nrt \ -- on the basis of \nrf. 
A given sentiment target in \nrf may be the target of multiple opinion expressions, each with different polarity and intensity. We assign a value from 1 to 3 to the sentiment intensities (`slight', `standard' or `strong'), and assign the sum of all sentiments to each target. This sum is clipped to a scale from -3 (strong negative) to 3 (strong positive) for each target. The data is made available on GitHub.\footnote{\url{https://github.com/ltgoslo/norec_tsa}}

\paragraph{Sentence-level sentiment}
We also derive a 4-class sentence-level sentiment dataset from \nrf. We take each sentence in \nrf with positive opinions only to be a positive sentence (and vice versa for negatives). Sentences without any opinion expressions are labeled "Neutral", and sentiments with both positive and negative sentiments are labeled "Mixed".

\paragraph{Document-level sentiment}
The document-level sentiment polarity is derived from the rating provided by the review author. 
Table~\ref{table:norec_fine_stats_rate} shows the distribution of documents and sentences relative to these ratings for the subset of data included in the \nrf training split and in our exploratory ELSA dataset (as further detailed below). For the same two respective subsets of data, Table~\ref{table:norec_fine_stats_cat} shows the distribution of documents and sentences across different domains of reviews.  

As we can see, the dataset is highly unbalanced. The extreme ratings of 1 and 6 are rare, and are in our experiments therefore merged with their adjacent ratings. We arrive at sentiment polarity classifications for each document by labeling ratings 1 and 2  as negative, rating 5 and 6 as positive, and we here categorize ratings 3 and 4 as neutral polarity.

In sum, the ecosystem of annotations derived from \nrf provides us with a multi-domain dataset of full-text reviews annotated for sentiment at multiple levels of analysis; document-level, sentence-level and target-level. This allows for comparing different strategies for aggregating SA information from different levels as estimators of entity-level sentiment. In order to evaluate these strategies and information sources, the next section describes how we create an exploratory ELSA dataset, adding information about entity-level sentiment for a subset of the documents in \nrf.

\begin{table}[bt]
\smaller 
\begin{center}
\begin{tabular}{@{}lrrrr@{}}
\toprule
& \multicolumn{2}{c}{\nrf Train}  & \multicolumn{2}{c}{ELSA subset}\\
\cmidrule(lr){2-3}   \cmidrule(l){4-5} 
Rating  & Docs & Sents & Docs & Sents \\
\midrule
1  &  8  &  151  & 1 & 32\\
2  &  27  &  475  & 7 & 121\\
3  &  62  &  1345  & 4 & 98\\
4  &  91  &  2504  & 15 & 399\\
5  &  109  &  3225  & 19 & 586\\
6  &  30  &  934 & 4 & 109\\
\midrule
Total  &  327  &  8634  & 50 & 1345\\
\bottomrule
\end{tabular}
\end{center}
\caption[Ratings distribution]{Distribution of review ratings (1--6) in the full \nrf training split and our ELSA subset.}
\label{table:norec_fine_stats_rate}
\end{table}

\begin{table}[bt]
\smaller 
\begin{center}
\begin{tabular}{@{}lrrrr@{}}

\toprule
& \multicolumn{2}{c}{\nrf Train}  & \multicolumn{2}{c}{ELSA subset}\\
\cmidrule(lr){2-3}   \cmidrule(l){4-5} 
Category & Docs & Sents & Docs & Sents \\
 \midrule
games  &  16  &  445   & 2 & 62 \\
literature  &  35  &  877   & 5 & 148 \\
misc  &  1  &  36   & 1 & 36 \\
music  &  111  &  1915   & 13 & 246 \\
products  &  30  &  1753   & 6 & 298 \\
restaurants  &  6  &  290   & 1 & 44 \\
screen  &  118  &  2920  & 20 & 449 \\
sports  &  2  &  149 & & \\
stage  &  8  &  249  & 2 & 62 \\
\midrule
Total  &  327  &  8634  & 50 & 1345\\
\bottomrule
\end{tabular}
\end{center}
\caption[Domain distribution]{Distribution of sentences and documents across the multiple domains in the NoReC review corpus.}
\label{table:norec_fine_stats_cat}
\end{table}

\subsection{An exploratory dataset for ELSA}
\label{sec:exploratory}
To create an evaluation dataset for ELSA, 
we sample 50 documents at random from the 327 documents in the \nrf train split. There are in total 1345 sentences in this evaluation set.

As a pre-processing step, Named Entities are extracted from the texts using the Huggingface pipeline and the pretrained ScandiNER.\footnote{Available at \url{https://huggingface.co/saattrupdan/nbailab-base-ner-scandi}}
language model, fine-tuned for NER on all Nordic languages, where the Norwegian part of the training data is provided by the NorNE corpus  \cite{jorgensen-etal-2020-norne}. The reported scores for Norwegian NER are good, $\approx$91\% \fen. Since our goal is to identify sentiment towards volitional entities only, we keep only the entities with PER and ORG label. We find the NER model to perform well, with few or none missed entities. During manual inspection, spurious entities were deleted, and the few misclassifications were corrected.

\paragraph{Resolving coreference by substring matching}
A volitional entity may have several entity mentions in the text. 
In order to cluster entity mentions in a text, we  resolved the various mentions of a volitional entity by simple substring matching, and also check for the genitive marker \norex{-s} in Norwegian, such that both \norex{Jo Nesbø}  and \norex{Nesbøs} are resolved to the same entity. One of the few errors introduced by this approach was that \norex{Elisabeth I} was resolved to the same entity as \norex{Elisabeth II}. These errors were subsequently corrected.

\subsubsection{Manual ELSA labeling}
For each unique entity in the text that the entity mentions represent, we manually evaluate the documents' sentiment towards these entities into the categories "Positive", "Negative", or "Neutral". For entities that are targets of both positive and negative sentiment expressions, we consider what the document as a whole conveys. This sentiment labeling was performed by one reader, after several readings, when necessary. 
Figure~\ref{fig:john_wayne} shows a constructed example text, with the extracted entities, their entity mentions, and their manual sentiment classification. 

\begin{figure}
\smaller 
    \centering
\begin{tcolorbox}[]
I met \hllb{John Wayne} yesterday. We said hello on the street when he was taking his grandchild for a walk. \hllb{John} is such a nice guy. Nothing like \hllr{Clint}, who is still very handsome, but seems quite arrogant.
\tcblower
\begin{tabular}{l l}
John Wayne [John Wayne, John]& Positive\\
Clint [Clint]& Negative\\
\end{tabular}
\end{tcolorbox}
    \caption{Example of entity mentions in text as positive/negative sentiment targets, together with the overall entity-level sentiment labels.}
    \label{fig:john_wayne}
\end{figure}

We can conclude from reading the text that the sentiment towards \textit{John Wayne} is positive, while the sentiment towards \textit{Clint} is negative. There is one positive sentiment expression towards \textit{Clint}: \textit{handsome}, and one negative expression: \textit{arrogant}. As readers of the entire text, we perceive the overall sentiment towards \textit{Clint} to be negative.
The entity \textit{John Wayne} has two entity mentions in the text: \textit{John Wayne} and \textit{John}. The pronoun \textit{he} is not an entity mention, but an anaphor, coreferential with \textit{John Wayne}. The grandchild is not an entity because it is not named.

By manually assessing the sentiment expressed towards each entity in the 50 selected documents, we arrive at a dataset of volitional entities and the sentiment towards them at the document-level. The majority of the entities are neutral, as shown in Table~\ref{table:elsa_data_polarity_stats}. This distribution is in line with \citet{mitchell-etal-2013-open} who find that for their Spanish twitter dataset 24\% of the entities are positive targets, 16\% of the entities are negative targets, and 61\% of the entities are neutral.	

\begin{table}
\smaller 
\begin{center}
\begin{tabular}{@{}lrrrr@{}}
\toprule
& ORG & PER & \multicolumn{1}{c}{\#} & \multicolumn{1}{c}{\%}\\ 
\midrule
Pos & 8 & 76 & 84 & 30\% \\ 
Neg & 3 & 25 & 28 & 10\% \\ 
Neu & 36 & 131 & 167 & 60\% \\ 
\midrule
Total & 47 & 232 & 279 & 100\% \\ 
\bottomrule
\end{tabular}
\end{center}
\caption[ELSA dataset polarity distribution]{The sampled subset of 50 documents from \nrf contains the names of 279 volitional entities. The manual annotations assigned a neutral entity-level sentiment to the majority of these.}
\label{table:elsa_data_polarity_stats}
\end{table}

\begin{table}
\smaller 
\begin{center}
\begin{tabular}{@{}lrr@{}}
\toprule
Mentions per entity & Entities & Entity mentions \\
\midrule
1 & 188 & 188 \\
2 & 39 & 78 \\ 
3+ & 52 & 266 \\
\midrule
Total & 279 & 532 \\
\bottomrule
\end{tabular}
\end{center}
\caption[ELSA dataset mentions]{ The evaluation data contain  279 volitional entities with an average of 1.9 entity mentions per entity. 33\% of the entities have more than one mention.}
\label{table:elsa_data_mentions}
\end{table}

\section{Analysis of target-independent approaches to ELSA}\label{sec:analysis_doc_sent}
If all volitional entities mentioned in a positive text are the target of positive sentiment, the task would be limited to an overall sentiment classification for the text. This naïve approach serves as a baseline for further studies.  In the following, we compare our manually labeled entity-level polarities with those of the documents they appear in, as well as those of each sentence with a corresponding entity mention.

\subsection{ELSA polarity vs. document polarity}
As mentioned in Section~\ref{sec:dataset}, each document is assigned the overall polarity positive, neutral or negative, based on its review rating.

\begin{figure}[tb]
	\centering
 \includegraphics[width=0.9\columnwidth]{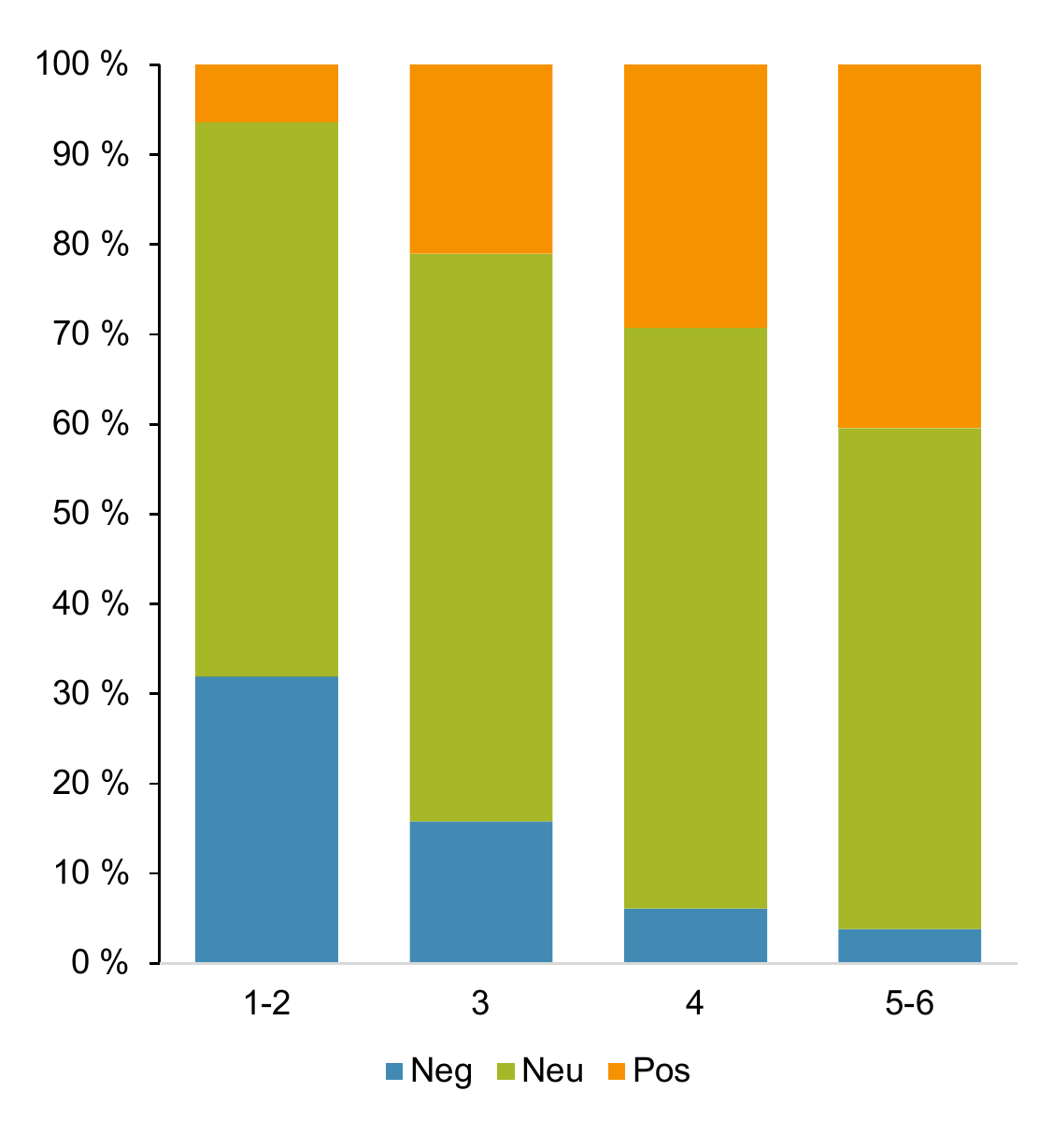}
	\caption{Relative distribution of entity polarities for the different document ratings. Neutral entities are in majority across all ratings.}
	\label{fig:doc_vs_elsa}

\end{figure}

In Figure~\ref{fig:doc_vs_elsa} and Table~\ref{table:doc_vs_elsa} we present the results of an analysis of entity polarities for the different document ratings. We find that only 47.7\% of the ELSA entities have the same polarity as the corresponding document-level label.  
We further observe that the neutral entities are quite evenly distributed -- between 55\% and 65\% across all document polarities. These results clearly suggest  that simply inferring entity-level polarity from the document-level is insufficient. 


\begin{table}[bt]
\smaller 
\begin{center}
    \begin{tabular}{@{}lrrrr@{}}
 \toprule
 &  \multicolumn{3}{c}{Entity polarities} & \\
  \cmidrule(lr){2-4}
 Rating& Neg & Neu & Pos & Entities \\
\midrule
1--2 & 15 & 29 & 3 & 47 \\
3 & 3 & 12 & 4 & 19 \\
4 & 5 & 53 & 24 & 82 \\
5--6 & 5 & 73 & 53 & 131 \\
\midrule
Total & 28 & 167 & 84 & 279\\
\multicolumn{2}{@{}l}{True pos}  & & &  133 \\
\multicolumn{2}{@{}l}{Accuracy}  & & &  0.477 \\
\bottomrule
\end{tabular}
\end{center}
\caption{Distribution of entity polarities for each of the document ratings categories. As ratings of 1 and 6 are rather scarce, we merge them with their adjacent rating.} 
\label{table:doc_vs_elsa}
\end{table}

\subsection{ELSA polarity vs sentence polarity}
We now turn to the sentence-level polarity labels presented in  Section~\ref{sec:dataset}.  
Since the ELSA entities may have multiple mentions, they may appear in multiple sentences. We here aggregate  the polarity towards an entity by considering it as positive if it is mentioned in more positive than negative sentences (and vice versa). The polarity is considered neutral if the entity appears in only neutral sentences. When entity mentions are equally present in positive and negative sentences, or in mixed polarity sentences only, mixed polarity is assigned.


\begin{figure}[tb]
	\centering
 \includegraphics[width=0.9\columnwidth]{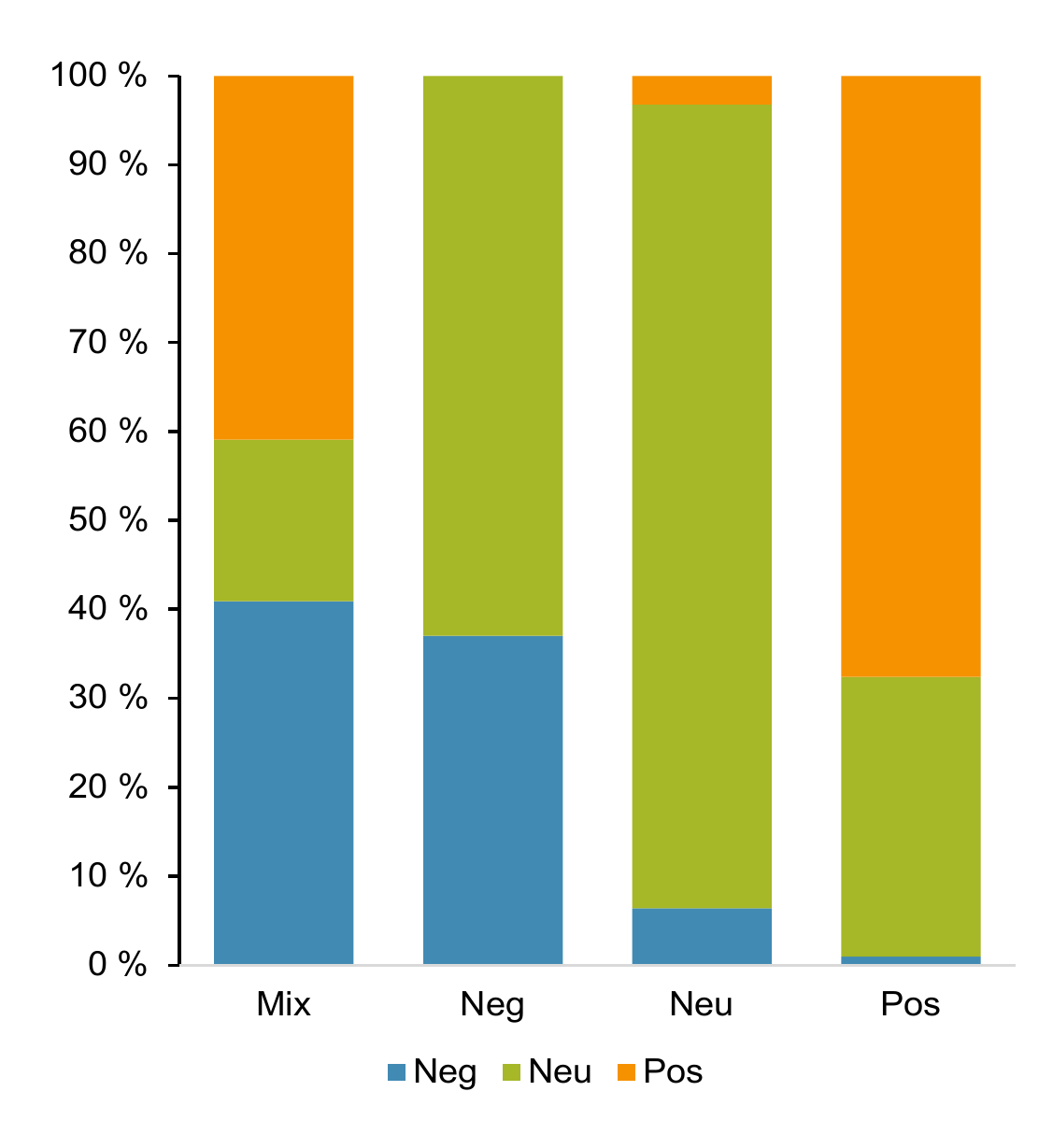}
	\caption{Relative distribution of entity polarities for each of the sentence sentiment categories.}
	\label{fig:sentence_vs_elsa}
\end{figure}

\begin{table}[bt]
\smaller 
\begin{center}
\begin{tabular}[l]{@{}lrrrr@{}}
    \toprule
     &  \multicolumn{3}{c}{Entity polarities} &\\
    \cmidrule(lr){2-4}
    Pol. & Neg & Neu & Pos & Tot. \\
    \midrule
    Mix &  9 & 4 & 9 & 22  \\
    Neg &  10 & 17 &  & 27 \\
    Neu &  8 & 113 & 4 & 125  \\
    Pos &  1 & 33 & 71 & 105  \\
    \midrule
    Total& 28 & 167 & 84 & 279  \\
    \multicolumn{2}{@{}l}{True pos}  & & &  194 \\
    \multicolumn{2}{@{}l}{Accuracy} & & & 0.695  \\

\bottomrule
\end{tabular}
\end{center}

\caption{Distribution of entity polarities for each of the sentence sentiment categories. Approximately 70\% of the entities were correctly labeled when using sentence polarity as a proxy.}

\label{table:sentence_vs_elsa}
\end{table}

The results are summarized in Figure~\ref{fig:sentence_vs_elsa} and Table~\ref{table:sentence_vs_elsa}. 
We find that neutral entities are more frequent than negative entities in sentences with negative polarity. An example of neutral entities that appear in a non-neutral sentence is provided in Example~\ref{ex:JA}. The sentence is classified as negative and without any sentiment targets. The annotated sentiment towards \textit{Julian Assange} is neutral.

\begin{exe}
\ex \label{ex:JA} Det gir en ganske merkelig effekt, litt som å treffe Julian Assange på Disneyland. \\
\textit{This has a quite peculiar effect, somewhat like meeting Julian Assange at Disneyland.}
 \end{exe}

116 of the entities (42\%), have mentions in the same sentence as an entity of a conflicting  polarity. The mixing of polarities inside a sentence indicates that inferring entity polarity from sentence polarity has limited potential. In sum, we find that 69.5\% of the entities would be correctly resolved by inferring entity-level sentiment from the sentence sentiment.

\section{A target-dependent approach to ELSA}\label{sec:analysis_target_dep}
An intuitively more promising approach is to derive ELSA sentiment labels from targeted sentiment analysis. 
Our TSA dataset presented in Section~\ref{sec:norec} is annotated for sentiment towards each target with a scale from -3 (strong negative) to 3 (strong positive). We aggregate these labels from the target-level to the entity-level for each entity by including only the sentiment targets that overlap with an entity mention, and summing the sentiment values for these targets. This approach leaves 7 entities with an unresolved classification due to equally strong positive and negative sentiment towards its entity mentions. These entities are placed in the "Mix" category.

\begin{figure}[tb]
	\centering
 \includegraphics[width=0.9\columnwidth]{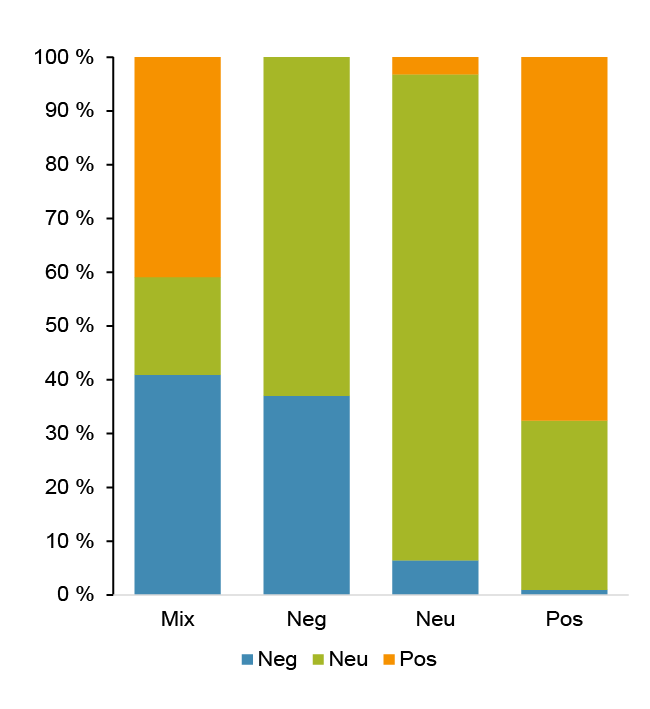}
	\caption{Relative distribution of entity polarities for each of the polarities derived from TSA.\\}
	\label{fig:tsa_vs_elsa}
\end{figure}

\begin{table}[bt]
\smaller 
\begin{center}
\begin{tabular}[c]{@{}lrrrr@{}}
\toprule
 &  \multicolumn{3}{c}{Entity polarities} & \\
\cmidrule(lr){2-4}
TSA & Neg & Neu & Pos &  Tot.\\
\midrule
Mix & 3 &  & 4 & 7 \\
Neg & 11 &  & 3 & 14  \\
Neu & 10 & 162 & 21 & 193   \\
Pos & 4 & 5 & 56 & 65  \\
\midrule
Total& 28 & 167 & 84 & 279  \\
\multicolumn{2}{@{}l}{True pos}  & & &  229 \\
\multicolumn{2}{@{}l}{Accuracy}  &  &  & 0.821\\
\bottomrule
\end{tabular}
\end{center}
\caption{Distribution of entity polarities for each of the polarities derived from TSA.}
\label{table:tsa_vs_elsa}
\end{table}

As shown in Table~\ref{table:tsa_vs_elsa}, 229 of our 279 volitional entities, 82\%, are given the correct sentiment label when aggregating the TSA annotations. This is an encouraging improvement over previous results. Virtually all neutral entities receive a neutral label through this approach. The 50 entities that are not correctly labeled using this approach, however, call for further analysis.

\subsection{Studying the remaining 50 entities}

In a further analysis step, we manually inspect the misclassified entities following TSA to look for common causes of errors. 

\paragraph{Annotator disagreement}  
Table~\ref{table:50_ents_categories} shows that human disagreement was the most important cause. For these instances, we have interpreted the sentences differently with respect to sentiment, than the original annotators of \nrf. This serves as an example of the subjectivity and one may say fragility of human sentiment annotations. This is also evident in the moderate inter-annotator agreement for the \nrf annotations, reported to be 73\% \fen for targets, when counting binary overlap \citep{ovrelid-etal-2020-fine}. 


\begin{table}[bt]
\smaller 
\begin{center}
\begin{tabular}{@{}lrr@{}}
\toprule
Error type &  \# &     \%  \\
\midrule
Missing CR            &      2 &    0.72\% \\
Missing TER           &      18 &    0.65\% \\
Anno. disagreement    &      24 &    8.60\% \\
Mixed polarity &       6 &   2.15\% \\ 
\bottomrule
\end{tabular}
\end{center}
\caption{Distribution of different error types when classifying entity sentiment based on aggregated TSA: Missing coreference resolution (CR), missing  target--entity resolution (TER), annotator disagreement, and cases of ties from mixed TSA polarities.}
\label{table:50_ents_categories}
\end{table}

\paragraph{Coreference resolution}
To our surprise, only two (out of 50) entities were incorrectly classified due to lack of coreference resolution. One such case is presented in example~(\ref{ex:JG}) below.

\begin{exe}
\ex \label{ex:JG}  Joseph Goebbels er på den måten i sjeldent, men ikke godt, selskap. Han er blitt stående som selve bildet på naziregimets hensynsløse og ondskapsfulle hatideologi. \\
\textit{Joseph Goebbels is in this respect in an exclusive, but not good company. He is the embodiment of the cruel and evil nazi regime's ideology of hatred.}
 \end{exe}

In Example~\ref{ex:JG}, \textit{Joseph Goebbels} in the first sentence is not annotated as a sentiment target in \nrf, while \norex{Han} \eng{He} in the second sentence is a negative sentiment target. Through coreference resolution, \norex{Han} \eng{He} would likely be resolved to refer to \textit{Joseph Goebbels}, and the sentiment towards the entity would be correctly resolved. 


\paragraph{Sentiment-relevant relation extraction} 
From our manual inspection we find another cause of misclassification pointing to the need for what we might call "target--entity resolution" rather than classical anaphora / cataphora coreference resolution. These are examples that require inference of semantic relations between different target expressions and  entities in the text. Typical examples of this in our data are examples of a work of art where the sentiment towards the work of art implies a sentiment towards the creator, or a noun describing a group of people where an entity is a member. 
 \begin{exe}
 \ex \label{ex:MB} Magnus Beite har skaffet filmen musikk som kler miljøet \\\textit{ 
 Magnus Beite has provided the film with music that suits the environment} \\

\ex \label{ex:beatles} Bandet bestående av Daniel Birkeland på gitar [\ldots] og trommis Helge Nyheim klarer virkelig å gjenskape Beatles [\ldots] på en måte som det står respekt av. \\
\textit{ The band consisting of Daniel Birkeland on guitar [\ldots] and drummer Helge Nyheim really manage to give life to  Beatles [\ldots] in a way that commands respect.} 

\end{exe}
 In Example~\ref{ex:MB}, \norex{musikk} \eng{music} is a positive sentiment target. When reading the full text, we also perceive this as a positive sentiment towards \norex{Magnus Beite}. In Example~\ref{ex:beatles}, \norex{Bandet} \eng{The band} is the sentiment target, and we perceive this to imply positive sentiment towards the individual members of the band as well.

For these types of examples, if we had access to a "member of"-relation between the band members and the band, and a "creator of"-relation between the photographer and the photograph, or more generally a relation of target--entity affiliation,   
the sentiment towards the volitional entities would have been resolved correctly. (Note that traditional aspect-categories would not be of help here, as we would still be missing the relations between the targets and the relevant entities.) These observations open new research questions, however, about which semantic relations are sentiment-relevant and under which circumstances.
\citet{jiang-etal-2011-target} also point to this question, with the example that  a sentiment about someone’s behavior usually means a sentiment about the person, while a sentiment about someone’s colleague usually has nothing to do with the person.

\section{Modeling}\label{sec:modeling} 
The above data analysis was performed using gold standard data for the various sentiment-levels (document, sentence and target). In order to gauge the performance attainable in a more realistic setting, we present results using automatically derived TSA information in the following.

Since we have no directly annotated ELSA training data, we create a baseline model using the proxy task of TSA. The exploratory ELSA dataset was taken from the training split of \nrf, and we therefore join the remainder of the training split with the development split to create our baseline training data, and we perform the final evaluation on the 50 documents in our ELSA dataset. The model setup is adapted from the Huggingface example configuration for NER.\footnote{\url{https://github.com/huggingface/transformers/tree/main/examples/pytorch/token-classification}}  
We use default hyperparameters and perform no hyperparameter tuning. We therefore chose to not set aside data for a dev set, and not touch the original \nrf test split, in order to allow for further research on this data split. We preprocess our training data the same way as we did with the ELSA data in Section~\ref{sec:analysis_target_dep},  finding PER and ORG labels in the data through NER, and deriving sentiment towards these through the pre-existing TSA target annotations. Volitional entities that are sentiment targets, receive the sentiment label "Positive" or "Negative", while the volitional entities that are not sentiment targets are labeled "Neutral".
With this setup, volitional entities and their sentiment polarity is predicted for the 50 documents in the exploratory dataset. The predicted entity mentions were resolved document-wise through substring matching, before summing the predicted polarities. For evaluation, these predictions are compared with our manually resolved and annotated entities.\footnote{Code for our data collection, analysis and modeling will be available at \url{https://github.com/egilron/elsa-introduction}} Our evaluation ignores whether the model assigned the correct NER category PER or ORG. Our evaluation shows that this baseline model has a \fen-score of 70.1\%, as shown in Table~\ref{table:cm}. 

\begin{table}
\smaller 
\centering 
\begin{tabular}[l]{@{}lrrrr@{}}
\toprule
     &  \multicolumn{3}{c}{Gold entity pol} &\\
     \cmidrule(lr){2-4}
Pred entity pol &   Neg &    Neu &   Pos &  FP \\
\midrule
Neg &   4 &    0 &   4 &   1 \\
Neu &  13 &  154 &  27 &  35 \\
Pos &   9 &    6 &  48 &   8 \\
Missed &   2 & 7 &   5 &    \\
\bottomrule
\end{tabular}
\caption{A confusion matrix of gold and predicated entity polarities. The final column indicates false positive entity predictions. Our simple NER+TSA-based baseline model for ELSA predicted the name and sentiment polarity correctly for 206 of the 279 entities in the test data. The model produced 44 false positive entities, resulting in Precision: 66.7\%, Recall: 73.8\% and \fen: 70.1\%.}
\label{table:cm}
\end{table} 

\section{Conclusion} \label{sec:conclusion}

In this paper we have explored the task of entity-level sentiment analysis (ELSA) -- the task of determining the aggregated or overall (i.e. document-level) 
polarity expressed towards an entity in a text. In particular for longer text, which might comprise several distinct entities, entity mentions and opinions, this is a potentially complex task. The paper has surveyed existing literature and terminology in adjacent and related previous work, in addition to presenting an exploratory data analysis to shed more light on what is required to solve the task. 

Relating to the latter point, in order to assess the relevance to ELSA of existing approaches to sentiment analysis at different granularities -- i.e. document-,  sentence-, and target-level SA -- we perform a task analysis on the basis of a Norwegian multi-domain review dataset containing all these layers of sentiment annotation. 
When utilizing the fine-grained sentiment annotations of the \nrf dataset, we found that the overlap between a volitional entity and TSA annotations, gave us the correct sentiment category for 82\% of our 279 manually evaluated ELSA  entities. We further found that, using automated TSA, we obtained an \fen-score of 70.1\%. TSA therefore appears to be highly relevant, though not sufficient, for the ELSA task.

For the ELSA classifications that could not be derived from NER and TSA annotations, we found surprisingly few cases that would be correctly resolved through coreference resolution. This is in line with the findings of \citet{de-clercq-hoste-2020-absolutely}.
Our dataset is likely too small, however, to draw any definitive conclusions about the importance of coreference resolution for ELSA. On the other hand, we did observe a need for more generally resolving relations that tie target expressions to their corresponding entities. 

In our exploratory data analysis, we have identified at least four categories of information that could potentially benefit the classification of a document's overall sentiment towards entities: (\textit{i}) named entity recognition, (\textit{ii}) targeted sentiment analysis, (\textit{iii}) coreference resolution, and (\textit{iv}) what we dub target--entity resolution.  
The latter being a concept derived from working with the examples in our dataset, 
and refers to the task of identifying which entity a given target expression relates to. 
This appears to be a missing link in a pipeline for ELSA based on TSA, and to explore methods for filling this gap is a suggestion for further work.  

\section*{Acknowledgements}
The work documented in this publication has been carried out within the NorwAI Centre for Research-based Innovation, funded by the Research Council of Norway (RCN), with grant number 309834, and the SANT project (Sentiment Analysis for Norwegian Text), also funded by the RCN (grant number 270908). 

\bibliographystyle{acl_natbib}
\bibliography{bibliography}

\begin{thebibliography}{21}
\expandafter\ifx\csname natexlab\endcsname\relax\def\natexlab#1{#1}\fi

\bibitem[{Alimova and Tutubalina(2019)}]{alimova-tutubalina-2019-entity}
Ilseyar Alimova and Elena Tutubalina. 2019.
\newblock \href {https://aclanthology.org/W19-3641} {Entity-level
  classification of adverse drug reactions: a comparison of neural network
  models}.
\newblock In \emph{Proceedings of the 2019 Workshop on Widening NLP}, pages
  132--134, Florence, Italy. Association for Computational Linguistics.

\bibitem[{Barnes et~al.(2022)Barnes, Oberl{\"a}nder, Troiano, Kutuzov,
  Buchmann, Agerri, {\O}vrelid, and Velldal}]{barnes-etal-2022-semeval}
Jeremy Barnes, Laura Ana~Maria Oberl{\"a}nder, Enrica Troiano, Andrey Kutuzov,
  Jan Buchmann, Rodrigo Agerri, Lilja {\O}vrelid, and Erik Velldal. 2022.
\newblock {S}em{E}val-2022 task 10: Structured sentiment analysis.
\newblock In \emph{Proceedings of the 16th International Workshop on Semantic
  Evaluation (SemEval-2022)}, Seattle. Association for Computational
  Linguistics.

\bibitem[{De~Clercq and Hoste(2020)}]{de-clercq-hoste-2020-absolutely}
Orphee De~Clercq and Veronique Hoste. 2020.
\newblock \href {https://aclanthology.org/2020.crac-1.2} {It{'}s absolutely
  divine! can fine-grained sentiment analysis benefit from coreference
  resolution?}
\newblock In \emph{Proceedings of the Third Workshop on Computational Models of
  Reference, Anaphora and Coreference}, pages 11--21, Barcelona, Spain
  (online). Association for Computational Linguistics.

\bibitem[{De~Clercq et~al.(2011)De~Clercq, Hoste, and
  Hendrickx}]{de-clercq-etal-2011-cross}
Orph{\'e}e De~Clercq, V{\'e}ronique Hoste, and Iris Hendrickx. 2011.
\newblock \href {https://aclanthology.org/R11-1026} {Cross-domain {D}utch
  coreference resolution}.
\newblock In \emph{Proceedings of the International Conference Recent Advances
  in Natural Language Processing 2011}, pages 186--193, Hissar, Bulgaria.
  Association for Computational Linguistics.

\bibitem[{Engonopoulos et~al.(2011)Engonopoulos, Lazaridou, Paliouras, and
  Chandrinos}]{engonopoulos2011els}
Nikos Engonopoulos, Angeliki Lazaridou, Georgios Paliouras, and Konstantinos
  Chandrinos. 2011.
\newblock {ELS}: a word-level method for entity-level sentiment analysis.
\newblock In \emph{Proceedings of the International Conference on Web
  Intelligence, Mining and Semantics (WIMS '11)}, pages 1--9, Sogndal, Norway.

\bibitem[{Farra and McKeown(2017)}]{farra-mckeown-2017-smarties}
Noura Farra and Kathy McKeown. 2017.
\newblock \href {https://aclanthology.org/E17-1094} {{SMART}ies: Sentiment
  models for {A}rabic target entities}.
\newblock In \emph{Proceedings of the 15th Conference of the {E}uropean Chapter
  of the Association for Computational Linguistics: Volume 1, Long Papers},
  pages 1002--1013, Valencia, Spain. Association for Computational Linguistics.

\bibitem[{Huang and Fang(2020)}]{huang2020entity}
Zhihong Huang and Zhijian Fang. 2020.
\newblock An entity-level sentiment analysis of financial text based on
  pre-trained language model.
\newblock In \emph{2020 IEEE 18th International Conference on Industrial
  Informatics (INDIN)}, volume~1, pages 391--396. IEEE.

\bibitem[{Jiang et~al.(2011)Jiang, Yu, Zhou, Liu, and
  Zhao}]{jiang-etal-2011-target}
Long Jiang, Mo~Yu, Ming Zhou, Xiaohua Liu, and Tiejun Zhao. 2011.
\newblock \href {https://aclanthology.org/P11-1016} {Target-dependent {T}witter
  sentiment classification}.
\newblock In \emph{Proceedings of the 49th Annual Meeting of the Association
  for Computational Linguistics: Human Language Technologies}, pages 151--160,
  Portland, Oregon, USA. Association for Computational Linguistics.

\bibitem[{J{\o}rgensen et~al.(2020)J{\o}rgensen, Aasmoe, Ruud~Husev{\aa}g,
  {\O}vrelid, and Velldal}]{jorgensen-etal-2020-norne}
Fredrik J{\o}rgensen, Tobias Aasmoe, Anne-Stine Ruud~Husev{\aa}g, Lilja
  {\O}vrelid, and Erik Velldal. 2020.
\newblock \href {https://aclanthology.org/2020.lrec-1.559} {{N}or{NE}:
  Annotating named entities for {N}orwegian}.
\newblock In \emph{Proceedings of the 12th Language Resources and Evaluation
  Conference}, pages 4547--4556, Marseille, France. European Language Resources
  Association.

\bibitem[{Lee et~al.(2013)Lee, Chang, Peirsman, Chambers, Surdeanu, and
  Jurafsky}]{lee2013deterministic}
Heeyoung Lee, Angel Chang, Yves Peirsman, Nathanael Chambers, Mihai Surdeanu,
  and Dan Jurafsky. 2013.
\newblock Deterministic coreference resolution based on entity-centric,
  precision-ranked rules.
\newblock \emph{Computational linguistics}, 39(4):885--916.

\bibitem[{Li and Lu(2017)}]{li2017learning}
Hao Li and Wei Lu. 2017.
\newblock Learning latent sentiment scopes for entity-level sentiment analysis.
\newblock In \emph{Proceedings of the Thirty-First AAAI Conference on
  Artificial Intelligence}, San Francisco, USA.

\bibitem[{Luo et~al.(2022)Luo, Cai, Yang, Qin, Xia, and
  Zhang}]{https://doi.org/10.48550/arxiv.2204.06893}
Yun Luo, Hongjie Cai, Linyi Yang, Yanxia Qin, Rui Xia, and Yue Zhang. 2022.
\newblock \href {https://doi.org/10.48550/ARXIV.2204.06893} {Challenges for
  open-domain targeted sentiment analysis}.

\bibitem[{Mitchell et~al.(2013)Mitchell, Aguilar, Wilson, and
  Van~Durme}]{mitchell-etal-2013-open}
Margaret Mitchell, Jacqui Aguilar, Theresa Wilson, and Benjamin Van~Durme.
  2013.
\newblock \href {https://aclanthology.org/D13-1171} {Open domain targeted
  sentiment}.
\newblock In \emph{Proceedings of the 2013 Conference on Empirical Methods in
  Natural Language Processing}, pages 1643--1654, Seattle, Washington, USA.
  Association for Computational Linguistics.

\bibitem[{{\O}vrelid et~al.(2020){\O}vrelid, M{\ae}hlum, Barnes, and
  Velldal}]{ovrelid-etal-2020-fine}
Lilja {\O}vrelid, Petter M{\ae}hlum, Jeremy Barnes, and Erik Velldal. 2020.
\newblock \href {https://aclanthology.org/2020.lrec-1.618} {A fine-grained
  sentiment dataset for {N}orwegian}.
\newblock In \emph{Proceedings of the 12th Language Resources and Evaluation
  Conference}, pages 5025--5033, Marseille, France. European Language Resources
  Association.

\bibitem[{Pontiki et~al.(2016)Pontiki, Galanis, Papageorgiou, Androutsopoulos,
  Manandhar, AL-Smadi, Al-Ayyoub, Zhao, Qin, De~Clercq, Hoste, Apidianaki,
  Tannier, Loukachevitch, Kotelnikov, Bel, Jim{\'e}nez-Zafra, and
  Eryi{\u{g}}it}]{pontiki-etal-2016-semeval}
Maria Pontiki, Dimitris Galanis, Haris Papageorgiou, Ion Androutsopoulos,
  Suresh Manandhar, Mohammad AL-Smadi, Mahmoud Al-Ayyoub, Yanyan Zhao, Bing
  Qin, Orph{\'e}e De~Clercq, V{\'e}ronique Hoste, Marianna Apidianaki, Xavier
  Tannier, Natalia Loukachevitch, Evgeniy Kotelnikov, Nuria Bel,
  Salud~Mar{\'\i}a Jim{\'e}nez-Zafra, and G{\"u}l{\c{s}}en Eryi{\u{g}}it. 2016.
\newblock \href {https://doi.org/10.18653/v1/S16-1002} {{S}em{E}val-2016 task
  5: Aspect based sentiment analysis}.
\newblock In \emph{Proceedings of the 10th International Workshop on Semantic
  Evaluation ({S}em{E}val-2016)}, pages 19--30, San Diego, California.
  Association for Computational Linguistics.

\bibitem[{Pontiki et~al.(2014)Pontiki, Galanis, Pavlopoulos, Papageorgiou,
  Androutsopoulos, and Manandhar}]{pontiki-etal-2014-semeval}
Maria Pontiki, Dimitris Galanis, John Pavlopoulos, Harris Papageorgiou, Ion
  Androutsopoulos, and Suresh Manandhar. 2014.
\newblock \href {https://doi.org/10.3115/v1/S14-2004} {{S}em{E}val-2014 task 4:
  Aspect based sentiment analysis}.
\newblock In \emph{Proceedings of the 8th International Workshop on Semantic
  Evaluation ({S}em{E}val 2014)}, pages 27--35, Dublin, Ireland. Association
  for Computational Linguistics.

\bibitem[{Steinberger et~al.(2017)Steinberger, Hegele, Tanev, and
  Della~Rocca}]{steinberger-etal-2017-large}
Ralf Steinberger, Stefanie Hegele, Hristo Tanev, and Leonida Della~Rocca. 2017.
\newblock \href {https://doi.org/10.26615/978-954-452-049-6_091} {Large-scale
  news entity sentiment analysis}.
\newblock In \emph{Proceedings of the International Conference Recent Advances
  in Natural Language Processing, {RANLP} 2017}, pages 707--715, Varna,
  Bulgaria. INCOMA Ltd.

\bibitem[{Stoyanov and Cardie(2006)}]{stoyanov-cardie-2006-partially}
Veselin Stoyanov and Claire Cardie. 2006.
\newblock \href {https://aclanthology.org/W06-1640} {Partially supervised
  coreference resolution for opinion summarization through structured rule
  learning}.
\newblock In \emph{Proceedings of the 2006 Conference on Empirical Methods in
  Natural Language Processing}, pages 336--344, Sydney, Australia. Association
  for Computational Linguistics.

\bibitem[{Sweeney and Padmanabhan(2017)}]{sweeney-padmanabhan-2017-multi}
Colm Sweeney and Deepak Padmanabhan. 2017.
\newblock \href {https://doi.org/10.26615/978-954-452-049-6_094} {Multi-entity
  sentiment analysis using entity-level feature extraction and word embeddings
  approach}.
\newblock In \emph{Proceedings of the International Conference Recent Advances
  in Natural Language Processing, {RANLP} 2017}, pages 733--740, Varna,
  Bulgaria. INCOMA Ltd.

\bibitem[{Velldal et~al.(2018)Velldal, {\O}vrelid, Bergem, Stadsnes, Touileb,
  and J{\o}rgensen}]{velldal-etal-2018-norec}
Erik Velldal, Lilja {\O}vrelid, Eivind~Alexander Bergem, Cathrine Stadsnes,
  Samia Touileb, and Fredrik J{\o}rgensen. 2018.
\newblock \href {https://aclanthology.org/L18-1661} {{N}o{R}e{C}: The
  {N}orwegian review corpus}.
\newblock In \emph{Proceedings of the Eleventh International Conference on
  Language Resources and Evaluation ({LREC} 2018)}, Miyazaki, Japan. European
  Language Resources Association (ELRA).

\bibitem[{Zhang et~al.(2015)Zhang, Zhang, and Vo}]{zhang-etal-2015-neural}
Meishan Zhang, Yue Zhang, and Duy-Tin Vo. 2015.
\newblock \href {https://doi.org/10.18653/v1/D15-1073} {Neural networks for
  open domain targeted sentiment}.
\newblock In \emph{Proceedings of the 2015 Conference on Empirical Methods in
  Natural Language Processing}, pages 612--621, Lisbon, Portugal. Association
  for Computational Linguistics.

\end{thebibliography}
\end{document}